\documentclass{article}
\usepackage{spconf,amsmath,graphicx,hyperref}
\usepackage{amsmath,amssymb, graphicx, epstopdf,cite,enumerate,booktabs, setspace, float,stfloats,bm, multirow, lscape, caption, subcaption, graphbox, soul, color, xcolor}
\usepackage{bbding}
\usepackage{amsthm}
\usepackage{mathtools}
\usepackage[normalem]{ulem}
\usepackage[linesnumbered,ruled]{algorithm2e}

\usepackage{color}

\title{
LLM-Driven Scenario-Aware Planning for Autonomous Driving
\vspace{-0.1in}
}
\name{He Li$^{1}$, Zhaowei Chen$^{1}$, Rui Gao$^{3}$, Guoliang Li$^{1}$, Qi Hao$^{3}$, Shuai Wang$^{2,\dagger}$, and Chengzhong Xu$^{1,\dagger}$ 
\thanks{This work was supported by the National Key R\&D Program of China (No. 2025YFE0204100), Science and Technology Development Fund of Macao S.A.R (FDCT) under number 0074/2025/AMJ, the National Natural Science Foundation of China (Grant No. 62371444), and the Shenzhen Science and Technology Program (Grant No. RCYX20231211090206005, JCYJ20241202124934046).
Corresponding author: Chengzhong Xu ({\tt\small 
czxu@um.edu.mo}) and Shuai Wang ({\tt\small s.wang@siat.ac.cn})}}
\address{$^{1}$University of Macau
$^{2}$Shenzhen Institutes of Advanced Technology, Chinese Academy of Sciences\\
$^{3}$Southern University of Science and Technology}

\begin{document}
\maketitle

\begin{abstract}
Hybrid planner switching framework (HPSF) for autonomous driving needs to reconcile high-speed driving efficiency with safe maneuvering in dense traffic.
Existing HPSF methods often fail to make reliable mode transitions or sustain efficient driving in congested environments, owing to heuristic scene recognition and low-frequency control updates.
To address the limitation, this paper proposes LAP, a large language model (LLM) driven, adaptive planning method, which switches between high-speed driving in low-complexity scenes and precise driving in high-complexity scenes, enabling high qualities of trajectory generation through confined gaps.
This is achieved by leveraging LLM for scene understanding and integrating its inference into the joint optimization of mode configuration and motion planning. The joint optimization is solved using tree-search model predictive control and alternating minimization.
We implement LAP by Python in Robot Operating System (ROS). 
High-fidelity simulation results show that the proposed LAP outperforms other benchmarks in terms of both driving time and success rate. 

\end{abstract}

\vspace{-0.05in}
\begin{keywords}
Scenario-aware, autonomous driving, large language model
\end{keywords}

\vspace{-0.1in}
\section{Introduction}
\vspace{-0.1in}
How to drive efficiently under various driving scenarios, e.g., dense traffic, highway driving, is the crucial issue for autonomous driving (AD)\cite{zhao2025survey,gonzalez2015review}. 
Conventional AD methods struggle with adaptability: some universally apply strict collision avoidance, which is overly conservative in many contexts \cite{zhang2020optimization,rosmann2017kinodynamic}, while others are designed for high-speed driving but assume static or obstacle-free environments \cite{becker2023model,arab2023motion}.
To address this issue, a promising solution is a hybrid planner switching framework (HPSF), which can flexibly adapt to multiple driving modes instead of a single scenario.

The main challenge to realizing HPSF is determining the appropriate timing for switching between driving modes. 
First, switching to an inappropriate mode may compromise driving efficiency or even safety. For example, engaging dense-traffic mode on highways reduces speed, while activating high-speed mode in congested traffic increases collision risk.
Second, efficient driving in dense traffic remains challenging.
Existing HPSF approaches struggle to align driving styles with scene demands due to reliance on simple heuristic scene-recognition methods \cite{stahl2019multilayer, wang2021game}. Moreover, low operating frequency ($<20$Hz) in dense traffic \cite{zhang2020optimization, zhang2022generalized} further restricts performance across diverse scenarios.

To fill this gap, this paper proposes an \textbf{L}LM-driven \textbf{A}daptive \textbf{P}lanner, termed as LAP. The LAP method integrates an LLM-driven mode switcher, which can infer the appropriate driving strategy in real time like a human driver with a driving knowledge database and scenario description.
Guided by the LLM, the system dynamically selects between two complementary modes: a fast driving (FD) mode that adopts active maneuvers to push the vehicle to its dynamic limit in regions with sparsely distributed obstacles, and a shape-aware (SA) mode that carefully navigates in regions with densely distributed obstacles. Combining situation awareness with context-dependent planning enables robust real-time performance in diverse scenarios.
Technically, we determine when to engage FD or SA by solving an LLM-guided joint optimization of style configuration and motion planning. The optimization is carried out using tree-search model predictive control (TMPC) and alternating minimization (AM).
We implement the LAP in Python as a Robot Operating System (ROS) package and verify its performance on the Carla simulation platform.
Results show that LAP effectively adapts to diverse scenarios, achieving a higher driving speed and more reliable overtaking performance than other benchmarks.

\vspace{-0.1in}
\section{Problem Formulation}\label{section2_1} 
\vspace{-0.1in}

\begin{figure}[t]
    \centering
    \includegraphics[width=1\linewidth]{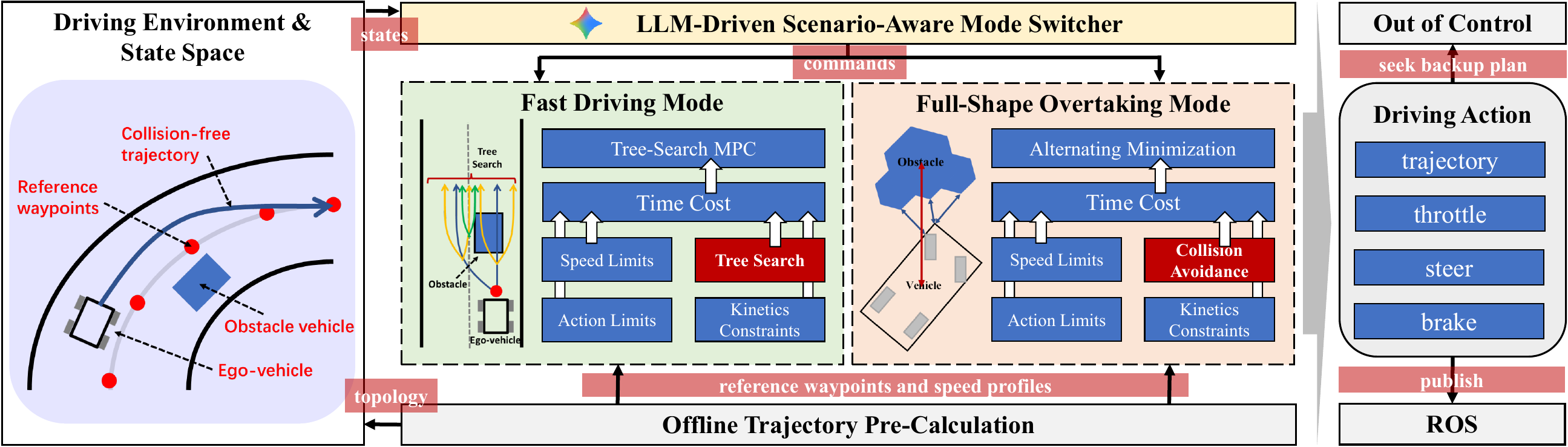}
    \caption{System architecture of LAP.}
    \label{fig:system}
    \vspace{-0.2in}
\end{figure}

We consider the driving scenario shown on the left-hand side of Fig.~\ref{fig:system}. 
We adopt the MPC formulation for HPSF, where the future states are predicted via robot dynamics over an $H$-step receding horizon.
It optimizes vehicle states $\{\mathbf{s}_t,\cdots,\mathbf{s}_{t+H-1}\}$ ($\mathbf{s}_t=[x_t,y_t,\theta_t]^T$ with position $(x_t,y_t)$ and orientation $\theta_t$) and actions $\{\mathbf{u}_t,\cdots, \mathbf{u}_{t+H-1}\}$ ($\mathbf{u}_t=[v_t,\psi_t]^T$ with linear velocity $v_t$ and angular velocity $\psi_t$)
under state evolution and action boundary constraints.
First, the state evolution is given by $\mathbf{s}_{t+1}=E(\mathbf{s}_{t},\mathbf{u}_{t})$, where
\begin{equation}
E\left(\mathbf{s}_{t},\mathbf{u}_{t}\right) = \mathbf{A}_{t}{\mathbf{s}_{t}} + \mathbf{B}_{t}{\mathbf{u}_{t}} + \mathbf{c}_{t},~\forall t,
  \label{dynamics}
\end{equation}
and $(\mathbf{A}_{t}$, $\mathbf{B}_{t}$, $\mathbf{c}_{t})$ are coefficient matrices of Ackermann kinetics defined in equations (8)--(10) of \cite[Sec. III-B]{han2023rda}.
Using \eqref{dynamics}, we can compute state $\mathbf{s}_{t+h+1}=
E(\mathbf{s}_{t+h},\mathbf{u}_{t+h})$ for any $h\in[0,H-1]$.
Second, the driving action is bounded by 
    \begin{align}
& \mathbf{u}_{\min } \preceq \mathbf{u}_{t+h} \preceq \mathbf{u}_{\max },~\forall h,  \nonumber
    \\
    &\mathbf{a}_{\min }  \preceq  {\mathbf{u}_{t+h+1}}
    -{\mathbf{u}_{t+h}}   \preceq \mathbf{a}_{\max },~\forall h, \label{bounds}
    \end{align}
where ${\mathbf{u}_{\min }}$ and ${\mathbf{u}_{\max }}$ are the minimum and maximum values of the control vector, respectively, and ${\mathbf{a}_{\min }}$ and ${\mathbf{a}_{\max }}$ are the associated minimum and maximum acceleration bounds.
The entire physical constraints are thus $\{ \mathbf{s}_{t+h}, \mathbf{u}_{t+h} \}\in\mathcal{P}$ with $\mathcal{P}=\{\{ \mathbf{s}_{t+h}, \mathbf{u}_{t+h} \}: (\ref{dynamics}), (\ref{bounds})\}$.
Given the pre-calculated pose ${\mathbf{s}_{t}^\diamond}$ and speed ${v_{t}^\diamond}$ (detailed in Section III-A), the HPSF problem $\mathsf{P}0$ is \cite{zhang2020optimization}:

\vspace{-0.2in}
\begin{subequations}\label{raw_problem}
\begin{align}
\mathsf{P}0\!:\!\!\!
\min_{\{ \mathbf{s}_{t+h}, \mathbf{u}_{t+h} \}\in\mathcal{P}}
\  &  \underbrace{\sum^{H}_{h=0} \Big( \|\mathbf{s}_{t+h} - \mathbf{s}_{t+h}^\diamond\|^2 
          + \|v_{t+h} - v_{t+h}^\diamond\|^2 \Big)}
          _{:=C_0\Big( \{ \mathbf{s}_{t+h}, \mathbf{u}_{t+h} \}_{h=0}^{H} \Big)}\label{raw_c0}\\
\text{s.t.} \ \ \ & {\bf{dist}}\left(\mathbf{s}_{t+h},\mathbf{o}_{i,t+h}\right) \geq d_{\mathrm{safe}}, 
  ~\forall i,h, \label{raw_const}
\end{align}
\end{subequations}
where \eqref{raw_const} is the collision avoidance constraint, ${\bf{dist}}\left(\mathbf{s}_{t},\mathbf{o}_{i,t}\right)$ is the distance between the ego-vehicle and the $i$-th obstacle at time $t$, and $d_{\mathrm{safe}}$ in meter is a pre-defined safety distance.

Solving problem $\mathsf{P}0$ is nontrivial due to the bilevel collision‑avoidance constraint \eqref{raw_const}. Embedding this constraint directly in the optimizer would compromise frequency, whereas approximating $\{\mathbf{o}_{i,t+h}\}$ into sets\cite{zhang2020optimization} or voxels\cite{zhang2020falco} would compromise accuracy. Below, we present an effective LLM solution to tackle this challenge.

\vspace{-0.1in}
\section{LLM Driven Adaptive Planning}\label{section3} 

\vspace{-0.1in}
\subsection{System Overview}

The architecture of LAP is shown in Fig.~\ref{fig:system}, which reads the topology and states including poses of ego- and other-vehicles (on the left hand side of Fig.~\ref{fig:system}) from the driving route and vehicle sensors, and outputs collision-free trajectories and associated driving actions including steer, throttle, and brake (on the right hand side of Fig.~\ref{fig:system}). 

The pipeline starts with offline pre-computation. 
The circuit is segmented into geometric primitives (straights/curves), and an optimization kernel pre-generates high-resolution, vehicle-dynamics-aware driving waypoints $\{\mathbf{s}_{t}^\diamond\}$ and speed envelopes $\{v_{t}^\diamond\}$ using an offline pure pursuit (PP) algorithm.
Note that the actual path may differ from the pre-calculated one due to dynamic obstacle avoidance.

During the online planning phase, the snapshot is fed to a scenario-aware planner switcher powered by an LLM. The switcher issues a command vector $\mathbf{c}=\{\,c_1,c_2\,\},$ where $c_1\in\{\mathsf{fd},\mathsf{sa}\}$ selects either the FD mode for aggressive behavior or the SA mode for conservative driving, and $c_2 \in\{\mathsf{acc},\mathsf{keep},\mathsf{dec}\}$ recommends a speed for acceleration or deceleration.
To translate text vector $\mathbf{c}$ into executable commands, we introduce mapping $\mathcal{M}_1:c_1\rightarrow \beta$ with $\beta\in\{0, 1\}$, where $\beta = 0$ selects FD mode, while $\beta = 1$ selects SA mode. 
To translate the speed reference vector, we introduce mapping 
$\mathcal{M}_2:c_2\rightarrow \gamma$, such that $\gamma \in \{v_0, 0, -v_0\}$, corresponding to $\{\mathsf{acc},\mathsf{keep},\mathsf{dec}\}$, respectively, where $v_0$ is the pre-defined velocity shift.

\vspace{-0.1in}
    \subsection{LAP Cost and Constraints} \label{sub_cost}

The cost function of LAP is based on \eqref{raw_c0} but needs to incorporate the LLM inference $\gamma$ for adaptive planning. 
Given the original reference speed  ${v_{t}^\diamond}$, the updated reference speed is  $v_{t}^\diamond \leftarrow {v_{t}^\diamond} + \gamma$. 
Then the cost function becomes
\begin{align}
    \phantom{C_1} & C_1\left(\{\mathbf{s}_{t+h}, \mathbf{u}_{t+h}\}_{h=0}^{H}\}\right)=
     \sum^{H}_{h=0} \Big(
    \left\|\mathbf{s}_{t+h}-\mathbf{s}_{t+h}^\diamond\right\|^2 \nonumber
    \\
    & \quad\quad\quad\quad\quad\quad\quad\quad\quad\quad+ \|v_{t+h} - v_{t+h}^\diamond-\gamma\|^2\Big),
\end{align}

On the other hand, LAP builds on \eqref{raw_const} and incorporates the mode $\beta$ to balance computational speed and obstacle-avoidance accuracy.
If $\beta=0$ (FD mode), 
we sample $M$ paths on the route, where the waypoints of the $m$-th path are $\mathcal{W}_m=\{\mathbf{w}_{m,1}, \mathbf{w}_{m,2},\cdots\}$. 
Then optimizing $\{\mathbf{s}_{t+h}\}$ is transformed into optimizing path selection $\alpha_m$ (with $\alpha_m\in\{0,1\}$ and $\sum_m\alpha_m=1$), and \eqref{raw_const} is reformulated as
\begin{align}
& (1-\beta) \left\|\mathbf{s}_{t+h}-\sum_{m=1}^M\alpha_m\mathbf{w}_{m,t+h}\right\| \leq \epsilon,~\forall h.
\nonumber\\
& (1-\beta) \!\left[\underbrace{
{\bf{dist}}\left(\sum_{m=1}^M\alpha_m\mathbf{w}_{m,t+h},\mathbf{o}_{i,t+h}\right)}_{:=\Theta(\{\alpha_m\})} \!- d_{\mathrm{safe}}\right]
 \!\!\geq 0,\forall i,h.  \label{collision2}
\end{align}
where $\epsilon>0$ is a small deviation constant.

If $\beta=1$ (SA mode), the LAP will consider the full shape of objects.
Accordingly, the minimum distance in \eqref{raw_const} between full-shape ego-vehicle and obstacle vehicle is given by 
\begin{align}
  \hspace{-1em}{\bf{dist}}(\mathbf{s}_t, \mathbf{o}_{i,t}) = 
  \min \left\{ {\left. {{{\left\| \mathbf{e} \right\|}_2}} \right|(\mathbb{Z}_t(\mathbf{s}_t) + \mathbf{e}) \cap \mathbb{O}(\mathbf{o}_{i,t}) \ne \emptyset } \right\}, 
\label{dist}
\end{align}
$\mathbb{Z}_t(\mathbf{s}_{t})$ and $\mathbb{O}(\mathbf{o}_{i,t})$ are the compact convex sets occupied by the ego and $i$-th obstacle vehicles, which can be represented by the conic inequality~\cite{boyd2004convex}:
\begin{align}
     &{\mathbb{O}_{i,t}} = \{ \mathbf{o} |\mathbf{D}_{i,t}\mathbf{o}{ \preceq }\mathbf{b}_{i,t}\}, \nonumber \\
     &\mathbb{Z}_t(\mathbf{s}_t) = \mathbf{R}_t(\mathbf{s}_t)\mathbf{z} + \mathbf{p}_t(\mathbf{s}_t),~\forall\mathbf{z}\in\mathbb{C}, 
  \\
  &\mathbb{C}= \{ \mathbf{z} | \mathbf{G}\mathbf{z}{ \preceq } \mathbf{h}\}, \nonumber
\end{align}
where, $[\mathbf{D}_{i,t}, \mathbf{b}_{i,t}]$ and $[\mathbf{G}, \mathbf{h}]$ are decided by the shape of $i$-th obstacle vehicle and ego-vehicle respectively. $\mathbf{p}_t(\mathbf{s}_t)$ and $\mathbf{R}_t(\mathbf{s}_t)$ are the transition and rotation matrices.  

The set computation in~\eqref{dist} is non-convex and non-differential. 
To resolve this challenge, we adopt the strong duality to transform constraint~\eqref{dist} into a more computationally efficient form~\cite{zhang2020optimization, han2023rda}, and then constraint~\eqref{raw_const} becomes
\begin{equation}
\begin{aligned}
& \beta
\begin{bmatrix}
-\bm\lambda_{i,t} \\
-\bm\mu_{i,t} \\
d_{\mathrm{safe}} - \bm\lambda_{i,t}^{\!\top}\mathbf D_{i,t}\mathbf p_t(\mathbf s_t)
   + \bm\lambda_{i,t}^{\!\top}\mathbf b_{i,t}
   + \bm\mu_{i,t}^{\!\top}\mathbf h \\
\bigl\| \mathbf D_{i,t}^{\!\top}\bm\lambda_{i,t}\bigr\|_* - 1      
\end{bmatrix}
\leq \mathbf 0, \\[4pt]
& \beta \left[\bm\mu_{i,t}^{\!\top}\mathbf G
+ \bm\lambda_{i,t}^{\!\top}\mathbf D_{i,t}\mathbf R_t(\mathbf s_t)\right]
= \mathbf{0},
\end{aligned}
\label{dual}
\end{equation}
where, $\bm{\lambda}_{i,t}$ and $\bm{\mu}_{i,t}$ are the dual variables.

\vspace{-0.1in}
\subsection{LLM-Driven LAP Algorithm Design}
Based on the analysis in \ref{sub_cost}, we write the problem $\mathsf{P}0$ as
\begin{flalign}
    \mathsf{P}1:~~~~
    &\min_{\mathclap{\substack{
        \{\alpha_m,\beta,\gamma\},\{\mathbf{s}_{t+h}, \mathbf{u}_{t+h}\}, \\ 
        \{\bm{\lambda}_{i,t+h}, \bm{\mu}_{i,t+h}\}}}}\quad
     ~~~~~~C_{2}\bigl(\{\mathbf{s}_{t+h}, \mathbf{u}_{t+h}\}_{h=0}^{H}\bigr)
    \\
    \hspace{-2em} \text{s.t.} ~~ 
    & \mathsf{constraints}~\eqref{dynamics},\eqref{bounds},\eqref{collision2},\eqref{dual} \nonumber
    \\ 
    & \hspace{-2em} \alpha_m, \beta\in\{0,1\},\ \gamma\in\{-v_0,0,v_0\},\ \sum_{m=1}^M\alpha_m=1. 
\end{flalign}
It can be seen that $\mathsf{P}1$ involves joint optimization of mode configuration and motion planning.

Conventional methods utilize a rule-based planner switcher with rules to select $\{\alpha_m,\beta,\gamma\}$.
However, pre-defined rules are difficult to generalize across diverse scenarios. To address this, we introduce an LLM-based switcher $\mathcal{L}$ (\emph{LLM-switcher}) that selects the appropriate driving mode. At each timestep $t$, the LLM receives as input the system state $\mathbf{s}_{t}$ and environmental context $\mathbf{p}_{t}$.
Following the retrieval-augmented generation (RAG) \cite{lewis2020retrieval} method, we create a database $\mathcal{U}$ with the retrieval mechanism $Q$, where $\mathcal{U}$ contains driving experience, including strategies that can be adopted when dealing with situations, and $Q$ uses $\mathbf{s}_{t}, \mathbf{p}_{t}$ to retrieve top $k$ historical experience $Q(\mathbf{s}_{t}, \mathbf{p}_{t}, \mathcal{U}) \rightarrow \{q_{t,1}, q_{t,2}, \dots, q_{t,k} \}$, that is used as reference control policy. Then the LLM generates the high-level commands $\mathbf{c}$ through chain-of-thought reasoning:
\begin{align}
    & \mathcal{L}(\mathbf{s}_{t}, \mathbf{p}_{t}, \{q_{t,1}, q_{t,2}, \dots, q_{t,k} \}) \rightarrow \mathbf{c}= \{c_1, c_2\}.
\end{align}

We then obtain the solution of $\{\beta,\gamma\}$ to $\mathsf{P}1$ as 
\begin{align}
    \beta=\mathcal{M}_1(c_1), \
    \gamma=\mathcal{M}_2(c_2).
\end{align}

With solution $\{\beta,\gamma\}$, we consider two cases. 
If $\beta = 0$ (FD mode), constraints \eqref{dynamics}, \eqref{bounds}, \eqref{collision2} become active, and $\mathsf{P}1$ is a mixed-integer nonlinear optimization problem.
To solve it, we first determine the path samples $\{\mathbf{w}_{m,t+h}\}$ using MPC under constraints \eqref{dynamics} and \eqref{bounds}. 
This would generate $M$ candidate solutions of $\{\mathbf{s}_{t+h}^{(m)},\mathbf{u}_{t+h}^{(m)}\}$.
Then we determine the optimal value of $\{\alpha_m\}$ and a tree search approach is adopted to explore all possible $\{\alpha_m\}$. 
For each fixed $\{\alpha_m\}$ we check if $\Theta(\{\alpha_m\})\geq d_{\mathrm{safe}}$ holds.
If so, we add this solution to the candidate set $\mathcal{F}=\mathcal{F}\bigcup \{\alpha_1,\alpha_2,\cdots\}$; otherwise this trail is abandoned.
The optimal $\{\alpha_1^*,\alpha_2^*,\cdots\}$ is selected from $\mathcal{F}$ by minimizing the deviation between the $\{\mathbf{s}_{t+h}^{(m)}\}$ and the target waypoints $\{\mathbf{s}^\diamond_{t+h}\}$. 
The optimal motion is $\{\mathbf{s}_{t+h}^*=\sum_m\alpha_m^*\mathbf{s}_{t+h}^{(m)},\mathbf{u}_{t+h}^*=\sum_m\alpha_m^*\mathbf{u}_{t+h}^{(m)}\}$.

If $\beta = 1$ (SA mode), constraints \eqref{dynamics}, \eqref{bounds}, and \eqref{dual} become active. This problem no longer involves integers $\{\alpha_m\}$ but is non-convex because of the coupled variables $\{\mathbf{s}_{t+h}, \mathbf{u}_{t+h}, \bm{\lambda}_{i,t+h}, \bm{\mu}_{i,t+h} \}$. 
Fortunately, the subproblem of $\{\mathbf{s}_{t+h}, \mathbf{u}_{t+h}\}$ is convex with given fixed $\{\bm{\lambda}_{i,t+h}, \bm{\mu}_{i,t+h}\}$; and vice versa. 
To this end, the SA mode leverages the AM framework to solve this optimization problem in an alternative way. 
The idea is to split the variables into two groups, $\{\mathbf{s}_{t+h}, \mathbf{u}_{t+h}\}$ and $\{\bm{\lambda}_{i,t+h}, \bm{\mu}_{i,t+h}\}$, and then update each group of variables with the other fixed. 
Each subproblem is efficiently solved by off-the-shelf software, e.g., cvxpy \cite{diamond2016cvxpy}. 
The AM procedure starts with an initial solution from the previous frame and terminates after $2$ to $3$ iterations. 
This completes a LAP iteration and we update $t\leftarrow t+1$. 
The LLM-switcher is periodically triggered at $1$\,Hz and reactively triggered upon obstacle detection. This trigger mechanism enables both continuous adaptation and safety-critical responses.
Note that motion control executes at up to $100$\,Hz.

\vspace{-0.15in}
\section{Experiments}
\vspace{-0.1in}

We implemented the proposed LAP system using Python in ROS.
The high-fidelity $\mathsf{MetaGrandPrix}$ simulation platform is used for evaluations, which is 1:1 constructed according to the real-world Macau Grand Prix circuit in the $\mathsf{CARLA}$ \cite{carla} simulator, available \href{https://github.com/MoCAM-ResearchGroup/grandprix}{here}.
Our LAP system is connected to $\mathsf{MetaGrandPrix}$ via ROS bridge\cite{carla_bridge}. 
All simulations are implemented on a Ubuntu workstation with an Intel Core i9-11900 CPU and an NVIDIA $4090$ GPU.

\begin{figure}[!t]
\vspace{-0.05in}
  \centering
  \begin{minipage}{0.49\linewidth}
    \centering
    \includegraphics[width=\linewidth]{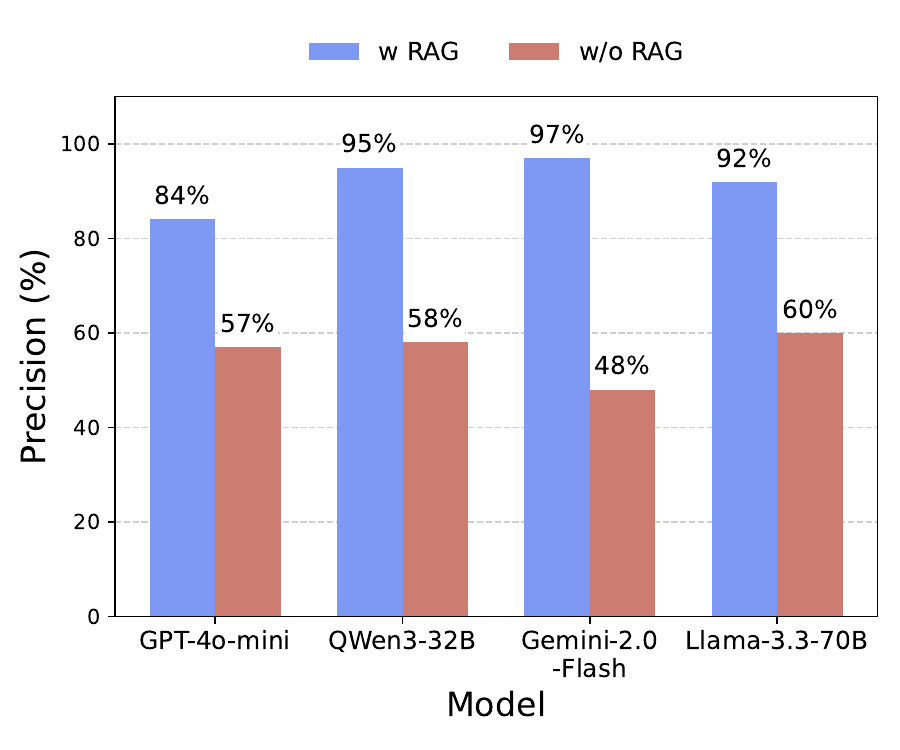}
    \vspace{-0.28in}
    \caption{Evaluation of LLMs.}
    \label{exp:1}
  \end{minipage}\hfill
  \begin{minipage}{0.49\linewidth}
    \centering
    \includegraphics[width=\linewidth]{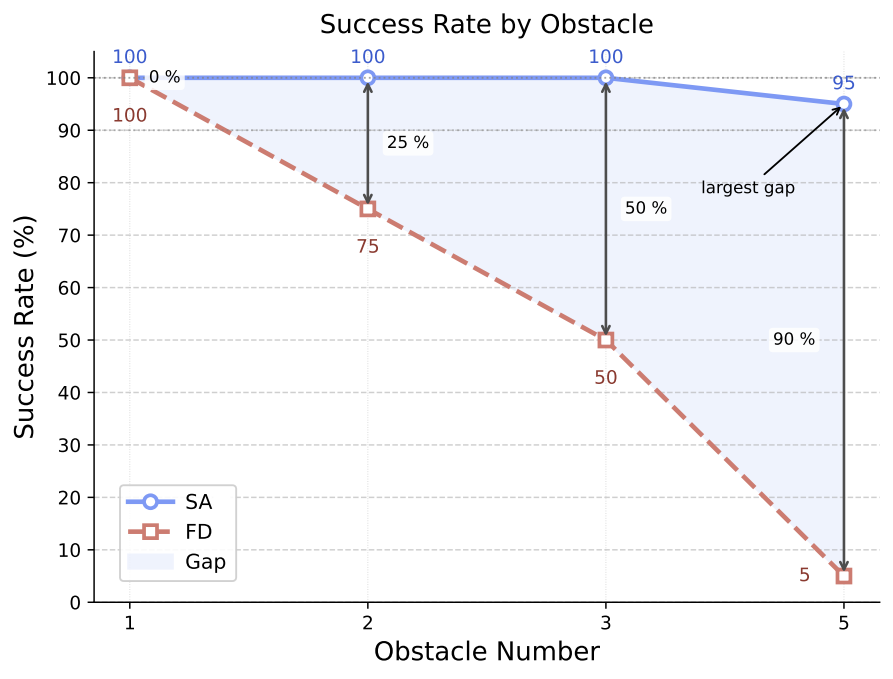}
    \caption{Overtaking results.}
    \label{exp:2}
  \end{minipage}
  \vspace{-0.05in}
\end{figure}

\begin{figure}[!t]
  \centering
  \begin{minipage}{0.35\linewidth}
    \centering
    \includegraphics[width=\linewidth]{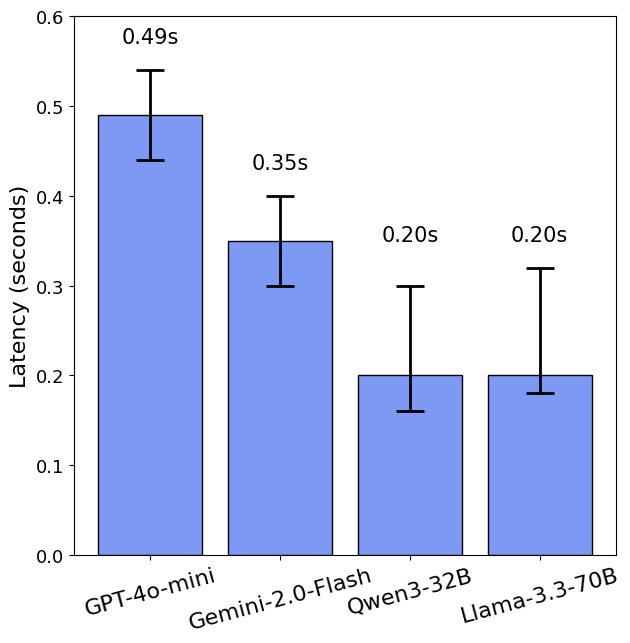}
    \vspace{-0.23in}
    \caption{TTFT.}
    \label{exp:llm_latency}
  \end{minipage}\hfill
  \begin{minipage}{0.64\linewidth}
    \centering
    \includegraphics[width=\linewidth]{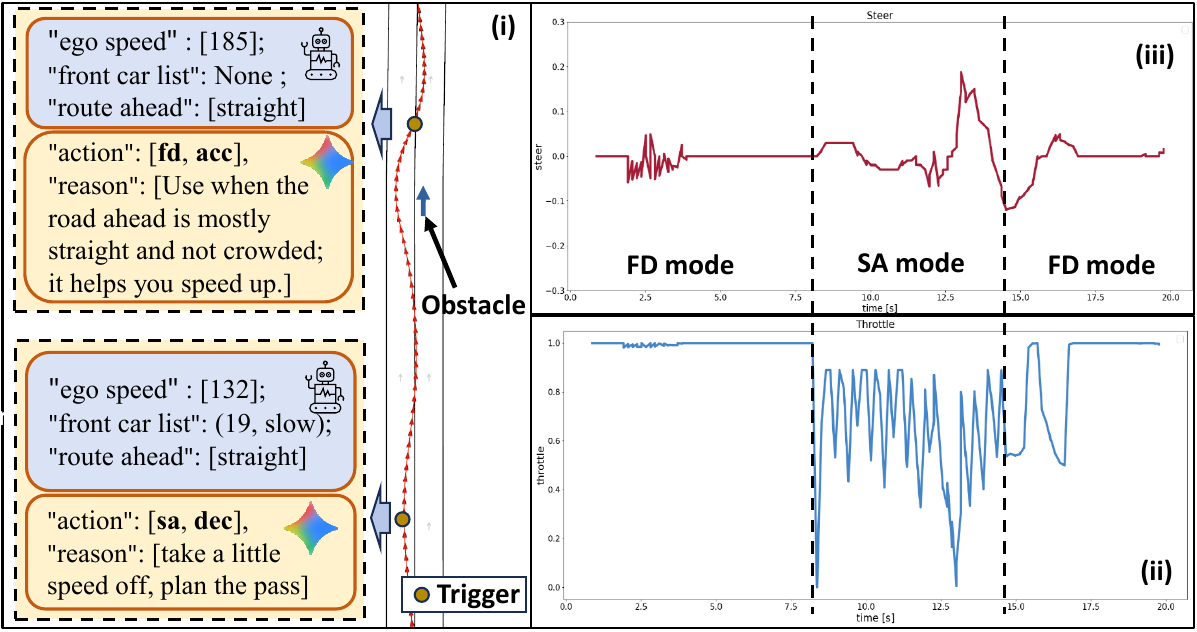}
    \caption{Illustration of LAP in CARLA.}
    \label{fig:demo}
  \end{minipage}
  \vspace{-0.05in}
\end{figure}

First, we evaluate the LLM-switcher across multiple models. 
We evaluate GPT-4o-mini \cite{openai2024gpt4o}, Qwen3-32B \cite{yang2025qwen3}, Gemini-2.0-Flash \cite{team2023gemini}, and Llama-3.3-70B \cite{dubey2024llama} as candidate LLMs. Given the real-time demands of driving, latency must be balanced with reasoning capability. As shown in Fig.~\ref{exp:llm_latency}, the average Time To First Token (TTFT) across all models is under 0.5\,s, indicating suitability for the LAP.
Specifically, we test candidate LLMs against an expert dataset comprising hundreds of samples, measuring precision as the match rate with expert outputs.
The results in Fig.~\ref{exp:1} show that RAG markedly improves the precision of all four LLMs in the scenario-aware switcher, with gains of at least 27\%. Gemini-2.0-Flash demonstrates the largest improvement, rising from 48\% without RAG to 97\% with RAG. Among RAG-enhanced models, Gemini-2.0-Flash outperforms GPT-4o-mini (84\%), Qwen3-32B (95\%), and Llama-3.3-70B (92\%). 
Therefore, we utilize Gemini-2.0-Flash with RAG as the mode switcher in the following experiments.

Next, we evaluate the driving capability of FD and SA modes under progressively denser traffic. They are required to perform overtake maneuvers with background vehicle density increases from one to five obstacles. Each mode was executed 20 times for each density level. Success rates are reported in Fig.~\ref{exp:2}.
LAP achieves 100\% success in the sparsest setting under both modes. As density increases, FD mode performance degrades sharply—75\% with two opponents, 50\% with three, and only 5\% with five. In contrast, SA mode remains robust, sustaining 100\% up to three opponents and 95\% at maximum density. These results show that FD mode suffices in light traffic, but the SA mode provides superior robustness in dense scenarios. 
The gap arises because FD models rivals as point centers and uses a coarse lane-based tree that collapses in congestion, whereas SA considers full vehicle geometry and preserves feasible corridors for overtaking.
This confirms the necessity of SA for high-density scenarios.

\begin{figure}[!t]
	\centering
	\begin{subfigure}{0.48\linewidth}
		\centering
		\includegraphics[width=\linewidth]{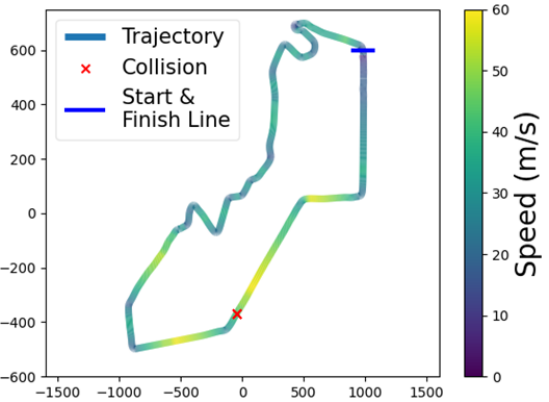}
		\caption{LAP.}
	\end{subfigure}
	\centering
	\begin{subfigure}{0.48\linewidth}
		\centering
		\includegraphics[width=\linewidth]{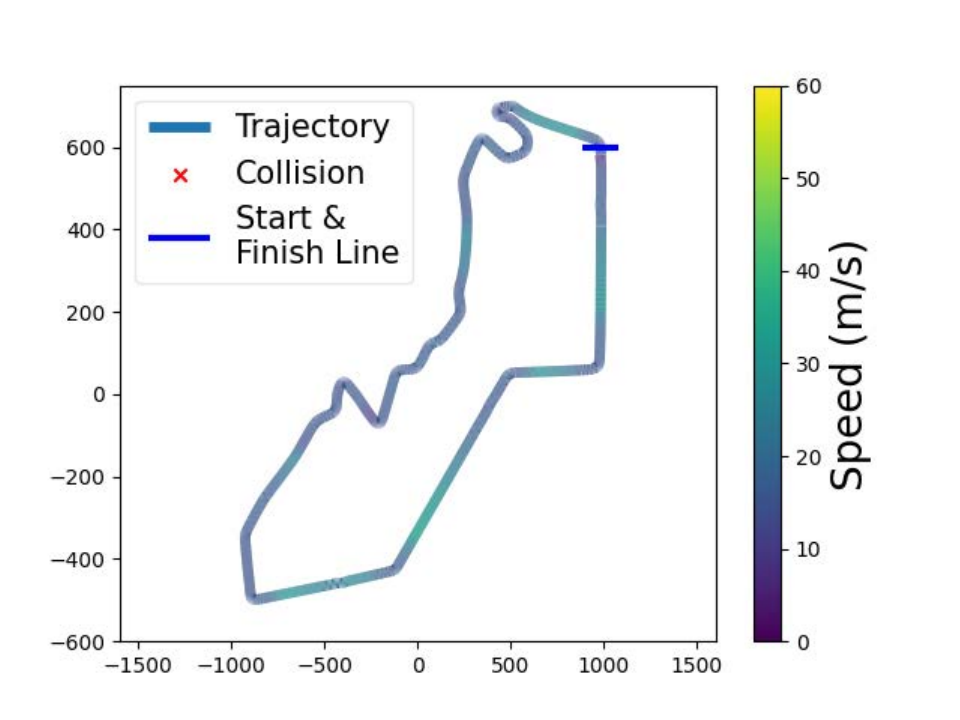}
		\caption{RDA.}
	\end{subfigure}
    \centering
    \vspace{-0.1in}
 	\caption{Racing trajectories.}
	\label{traj}
    \vspace{-0.2in}
\end{figure}

\begin{table}[!t]
  \centering
  \caption{Comparison of different racing approaches}
  \vspace{-0.05in}
  \label{result}
  \resizebox{\linewidth}{!}{
  \begin{tabular}{ccccc}
    \hline
    \multirow{2}{*}{Method} & lap time & avg.\ speed & colli.\ num & max speed\\
                            & (s)      & (km/h)      &      \dag       & (km/h)   \\ \hline
    CF-based PP                        & 932.70 & 25.09 & 12 & 51.05\\
    RDA \cite{han2023rda}              & 526.35 & 44.46 & 0 & 112.68\\
    FSM Speed\cite{meta}*              & 271.30 & 86.25 & 1  & --\\
    Optimistic \textbf{(Ours)}         & 265.65 & 88.09 & 1  & 211.31\\
    LAP \textbf{(Ours)}          & 269.13 & 86.95 & 1  & 211.20\\ \hline
  \end{tabular}
  }
  \caption*{\footnotesize * Champion of the 2023 Grand Prix Metaverse AD Challenge.\\
  \dag \ Racing allows collision as long as the vehicle can proceed.}
  \vspace{-0.35in}
\end{table}

Finally, we evaluate LAP in a high-speed scenario, measuring the best lap time over 10 trials. The track contains 10 sparsely distributed background vehicles controlled by a conventional MPC-based planner. LAP is compared against the following benchmarks: 
1) \textbf{CF-based PP:} CARLA’s built-in planner that follows the lead vehicle and accelerates when the lane is clear;
2) \textbf{RDA planner:} An accelerated full-shape MPC planner solving $\mathsf{P}1$ in parallel via ADMM \cite{han2023rda};
3) \textbf{Optimistic:} LAP without obstacles.
An illustration of LAP is shown in Fig.~\ref{fig:demo}, and the quantitative results are presented in Table.~\ref{result}.
The CF-based PP requires 932.7 s to complete the race (25 km/h average) due to its passive collision-avoidance strategy and poor lateral control. The shape-aware RDA planner halves the lap time to 526 s and eliminates collisions, but its operation frequency (i.e., 20 Hz) limits the peak speed at 113 km/h (as illustrated in Fig.~\ref{traj}b). In contrast, the proposed LAP finishes in 269.13 s, with an average speed of 86.9 km/h and peak velocity of 211 km/h. The racing trajectory of LAP in Fig.~\ref{traj}a also corroborates the performance. 
This performance is close to the optimistic upper bound under no obstacles, which is 265.65 s with a peak speed of 211 km/h.
Notably, the proposed LAP even outperforms the champion in the competition (271.3\,s)\cite{meta}, 
This result shows that the LLM-driven scenario-aware mode switcher can efficiently select the proper planner mode, and achieve the optimal balance between precise overtaking and fast driving.

\vspace{-0.15in}
\section{Conclusion}
\vspace{-0.1in}

This paper presented LAP, an HPSF method designed for dynamic driving scenarios. Experiments demonstrated that LAP balances the speed performance and safety assurance in dense traffic. In the high-speed setting, LAP switched between TMPC-based FD and AM-based SA, with TMPC driving the vehicle near its limits and AM enabling accurate overtaking. 
The vehicle with LAP was able to reach speeds of up to $211\,$km/h and achieve a lap time close to the optimistic scheme ($<5\,$s gap).

\bibliographystyle{IEEEbib}
\bibliography{refs}

\end{document}